\documentclass{article}

    \usepackage[nonatbib,preprint]{neurips_2025}

\usepackage[utf8]{inputenc} % allow utf-8 input
\usepackage[T1]{fontenc}    % use 8-bit T1 fonts
\usepackage{hyperref}       % hyperlinks
\usepackage{url}            % simple URL typesetting
\usepackage{booktabs}       % professional-quality tables
\usepackage{amsfonts}       % blackboard math symbols
\usepackage{nicefrac}       % compact symbols for 1/2, etc.
\usepackage{microtype}      % microtypography
\usepackage{xcolor}         % colors

\usepackage{wrapfig}

\usepackage{biblatex}
\usepackage[dvipsnames,svgnames, table,xcdraw]{xcolor}
\usepackage{relsize}
\usepackage[most]{tcolorbox}
\usepackage{listings}
\usepackage{subcaption}
\usepackage{algorithm}
\usepackage{algorithmic}
\usepackage{comment}
\usepackage{graphicx}
\usepackage{booktabs}
\usepackage{multirow}
\usepackage{enumitem}
\usepackage{url}
\usepackage{mathpazo}
\usepackage{amsmath,amssymb,amsthm,bm}
\usepackage[toc,page,header]{appendix}
\usepackage{fancyhdr}
\usepackage[T1]{fontenc}
\usepackage{cleveref}
\usepackage{adjustbox}
\usepackage{longtable}

\counterwithout{figure}{section}
\counterwithout{table}{section}
\definecolor{darkgoldenrod}{rgb}{0.72, 0.53, 0.04}
\definecolor{backgroundcolor}{RGB}{250, 250, 252}   
\definecolor{keywordcolor}{RGB}{30, 0, 178}       
\definecolor{stringcolor}{RGB}{204, 0, 102}        
\definecolor{numbercolor}{RGB}{0, 128, 128}        
\definecolor{emphcolor}{RGB}{30, 0, 178}            
\definecolor{commentcolor}{RGB}{0, 128, 0}       
\definecolor{basiccodecolor}{RGB}{61, 61, 61}

\lstdefinestyle{customstyle}{
    backgroundcolor=\color{backgroundcolor},   
    commentstyle=\color{commentcolor},
    keywordstyle=\color{keywordcolor},
    numberstyle=\color{numbercolor},
    stringstyle=\color{stringcolor},
    basicstyle=\color{basiccodecolor}\ttfamily\footnotesize,
    breakatwhitespace=false,         
    breaklines=true,                 
    captionpos=b,                    
    keepspaces=true,                 
    numbers=left,     
    basicstyle=\color{basiccodecolor}\ttfamily\footnotesize,
    numbersep=5pt,             
    xleftmargin=2em,
    xrightmargin=2em,
    showspaces=false,                
    showstringspaces=false,
    showtabs=false,                  
    tabsize=1,
    frame=single,
    framesep=5pt,
    framexleftmargin=1.5em,
    framexrightmargin=1.5em,
    framextopmargin=1pt,
    framexbottommargin=1pt,
    aboveskip=10pt,
    belowskip=10pt,
    breaklines=true,
    breakautoindent=true,
    emph={textgrad, tg, Variable, MultipleChoiceTestTime,
    TextualGradientDescent, BlackboxLLM},             %
    emphstyle={\color{emphcolor}},
    extendedchars=true,
}

\definecolor{logocolor}{RGB}{30, 0, 178}

\definecolor{darkerlogocolor}{RGB}{20, 0, 145}  

\newtcolorbox{ttcolorbox}[1][]{colframe=darkerlogocolor, colback=darkerlogocolor!4!white, title=#1}

\newtcolorbox{apxtcolorbox}[1][]{colframe=black, colback=black!3!white, title=#1}

\definecolor{ForestGreen}{RGB}{34,139,34} 
\definecolor{RoyalBlue}{RGB}{65,105,225}
\definecolor{TitleColor}{HTML}{B7B2D0}
\definecolor{ContentColor}{HTML}{DBD8E7}

\if false
\newcommand{\platform}[1]{ChatBattery}

\fi

\title{
LacMaterial: Large Language Models as Analogical Chemists for %Battery 
Materials Discovery 
}
\author{%
  Hongyu Guo %\thanks{Use footnote for providing further information     about author (webpage, alternative address)---\emph{not} for acknowledging     funding agencies.} \\
\\
National Research Council Canada\\
\url{https://hongyuharryguo.github.io}
}
\begin{document}

\maketitle

\begin{abstract}

Analogical reasoning, the transfer of relational structures across contexts (e.g., planet is to sun as electron is to nucleus), is fundamental to scientific discovery. Yet human insight is often constrained by domain expertise and surface-level biases, limiting access to deeper, structure-driven analogies both within and across disciplines. Large language models (LLMs), trained on vast cross-domain data, present a promising yet underexplored tool for  analogical reasoning in science. 
Here, we demonstrate that LLMs can generate novel battery materials by (1) retrieving cross-domain analogs and analogy-guided exemplars to steer  exploration beyond conventional dopant substitutions, and (2) constructing in-domain analogical templates from few labeled examples to guide targeted exploitation. These explicit analogical reasoning strategies yield candidates outside established compositional spaces and outperform standard prompting baselines. Our findings position LLMs as interpretable, expert-like hypothesis generators that leverage 
analogy-driven generalization for  
scientific innovation.

\end{abstract}

\begin{wrapfigure}{r}{0.35\textwidth}
\vspace{-6ex}
    \centering
    \includegraphics[width=0.851\linewidth]
    {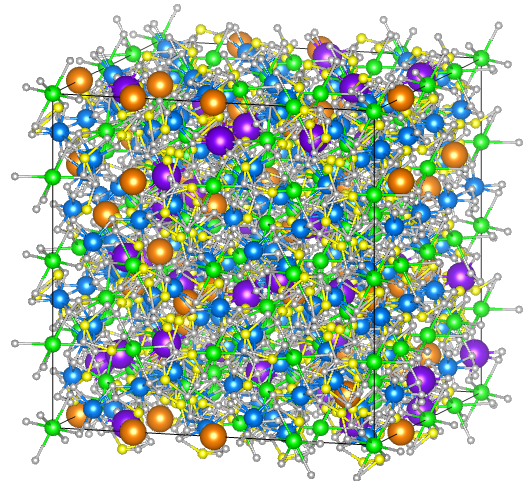}
    \vspace{-2ex}
   \caption{\small   {
    Supercell structure of the newly discovered  electrolyte Li$_{7.0}$La$_{2.5}$Nd$_{0.5}$Zr$_{1.4}$Hf$_{0.6}$O$_{12}$ } % (space group Ia-3d). 
   (La:blue; Nd:purple; Zr:yellow; Hf:orange; Li:green; O:gray).     }
    \label{fig:01_pipeline_crossdomain_supercell}
%\end{figure}
\vspace{-3ex}
\end{wrapfigure}
\section{Introduction}

Analogical reasoning, the ability to draw on prior knowledge to solve novel problems, is a hallmark of human intelligence and a foundational mechanism in scientific discovery~\cite{gentner1983structure,gentner2001analogical}. From ancient philosophical insights to modern breakthroughs, analogies have enabled scientists to generalize from familiar systems to unexplored domains.  For example, Maxwell used fluid and mechanical analogies to develop his theory of electromagnetism~\cite{harman1998maxwell}, while Luria likened bacterial mutation to slot machines to emphasize its random nature~\cite{luria1984slot}.

In materials science, analogical reasoning plays a similarly critical role. Domain experts routinely extrapolate structural and functional principles such as ion-conducting frameworks or dopant stabilization strategies from known compounds to guide the design of new materials~\cite{carter2013ceramic,goodenough2010challenges}. However, human analogical retrieval is often constrained by a bias toward surface-level similarity and limited exposure to cross-domain examples. This leads to a systematic preference for ``near” analogies within familiar domains, while ``far” analogies, those that are structurally relevant but perceptually less salient, are frequently overlooked~\cite{gentner1986systematicity,gick1983schema}. 
Large language models (LLMs), trained on %vast 
cross-domain data, present a promising yet underexplored tool for  addressing this limitation.

Recent advances show   LLMs exhibit emergent analogical reasoning abilities \cite{webb2023emergent}, yet these capabilities remain fragile:  highly dependent on prompt design and lacking systematic structure \cite{lewis2024format,mitchell2024fragile,yiu2025kiva}. Incorporating explicit, structured analogical frameworks markedly enhances %both
reasoning robustness% and interpretability
\cite{li2024analogykb,yasunaga2023analogical}, but their full transformative potential in driving scientific innovation remains largely untapped. 
Leveraging explicit structured analogical reasoning could elevate LLMs from passive pattern matchers to proactive engines of discovery, embracing selective exploitation of novel hypotheses. This raises a crucial question:
%\begin{quote}
\textit{How can LLMs harness explicit analogical reasoning to accelerate and steer scientific innovation?}
%\end{quote}

Consider, for example, an analogy from network engineering: in internet backbone networks, overlapping communication paths ensure robustness by avoiding single points of failure. Can we analogously design resilient solid electrolytes that support multiple, independent Li-ion migration pathways, so that localized disruptions (e.g., from dopant clustering or disorder) do not compromise long-range transport? This selective  shift in perspective, from optimizing  single-ion mobility to engineering percolation robustness through structural and chemical redundancy, reflects the power of deep analogy in   materials design. 

We here show that %, even without step-by-step prompts,
LLMs can effectively derive  both cross-domain and in-domain analogies to generate novel battery materials. Specifically, LLMs can 
 form cross-domain analogies to retrieve and adapt \textit{analogy-informed exemplars and structural motifs} from outside the target domain to inspire new material classes for exploration. Additionally, LLMs can  construct interpretable analogical rules and templates from a small set of labeled compounds, reflective of realistic research settings, for exploitation,  extrapolating from known materials to plausible candidates. 
 Our computational validation shows that these explicit analogical strategies %significantly
 broaden the design space beyond naive prompt-based generation. 
 \if false
{
%Promisingly
Interestingly, the LLM proposes %, to our knowledge, the first
a 
dual‑site, vacancy‑locked LLZO baseline explicitly framed to enable \emph{a priori} disentanglement of lattice vs.\ vacancy effects (Figure~\ref{fig:01_pipeline_crossdomain_supercell}), providing a clean, high‑value target for causal electrolyte design~\cite{Awaka2009,Buschmann2011,Rettenwander2016,Thangadurai2014}. }
\fi 
These findings suggest that explicit analogical reasoning enables LLMs to serve as transparent, expert-like hypothesis generators, navigating complex chemical spaces with both creativity and scientific grounding.

The remainder of the paper is organized as follows: Section~\ref{relatedwork} introduces related works. Section~\ref{exploration} describes how LLMs employ cross-domain analogical reasoning for materials exploration. Section~\ref{exploitation} presents how LLMs leverage in-domain analogies for materials exploitation. Finally, Section~\ref{conclusion} discusses  implications and outlines  future research.

\vspace{-1ex}
\section{Related Work} 
\label{relatedwork}
\vspace{-1ex}

Large language models (LLMs) have shown promising analogical reasoning abilities, which are essential for scientific discovery. While studies such as \cite{webb2023emergent} demonstrate LLMs’ success on analogy benchmarks, subsequent analyses \cite{lewis2024format,mitchell2024fragile} reveal that this reasoning is often fragile and reliant on superficial cues. Recent approaches to bolster LLM analogy robustness include  self-supervised analogy training \cite{zhou2025sal},  structured knowledge bases \cite{li2024analogykb}, and generating relevant exemplars for %in-context
analogical prompting \cite{%liu2025chatbattery,liu2023ChatDrug,
yasunaga2023analogical}. In ~\cite{yasunaga2023analogical}, LLMs are prompted to recall relevant cases for in-context learning, but it depends on the model's internal notion of relevance, which may conflate superficial resemblance with genuine relational similarity. 

Additionally, the above approaches primarily target abstract tasks rather than scientific domains. 
Identifying meaningful analogs in real-world scientific fields, particularly in materials design where inputs may be as sparse as a chemical formula, remains challenging. For example, 
Liu et al. \cite{liu2025chatbattery,liu2023ChatDrug} locate relevant   exemplars for in-context prompt based on  pre-defined, surface formula-level features. 

In contrast, our cross-domain analogy framework explicitly defines the analogy category for exemplar retrieval, ensuring source and target systems share deep relational structures rather than just surface-level features. Importantly, we constrain the LLM to reason within the selected analogy type, avoiding the open-ended and unfocused reasoning of prior work. 

In materials science, systems like ChatBattery \cite{liu2025chatbattery} use expert-guided prompting but lack explicit analogical reasoning frameworks. We fill this gap by introducing explicit structured, analogy-driven templates, making analogy a systematic tool for materials discovery.

\section{Cross-Domain Analogical Reasoning for Materials Generation} % for  \textit{Exploration}}
\label{exploration}
\vspace{-1ex}

Cross-domain analogical reasoning leverages abstract structural similarities across unrelated domains to inspire novel solutions. A classic example is the analogy between the solar system and the atom: despite surface-level differences (e.g., scale, color), both systems exhibit a central mass (sun, nucleus) orbited by smaller bodies (planets, electrons), revealing a shared relational structure. Such deep analogies are often difficult for humans to retrieve due to domain-specific biases and limited exposure.

Motivated by LLMs’ cross-domain training, we here aim to explore how LLMs can retrieve and apply explicit cross-domain analogies to materials discovery. Specifically, we prompt the LLM\footnote{All experiments used OpenAI O3, July 6: \url{https://openai.com/}} to generate analogical insights from its vast training corpus and use them to guide the design of novel solid-state electrolytes,  materials essential for next-generation batteries that enables  safer, higher-energy batteries than current liquid-based systems~\cite{goodenough2010challenges,janek2016solid}. 

Our focus is on cubic garnet LLZO (Li$_7$La$_3$Zr$2$O${12}$), a leading candidate among solid-state electrolytes for next-generation electric vehicles and grid-scale energy storage~\cite{janek2016solid,murugan2007fast}. 
LLZO offers exceptional electrochemical stability and resistance to air and moisture. 
In this study, we explore whether guiding LLMs via cross-domain analogical reasoning can inspire structural and chemical innovations for new LLZO-like electrolytes, pushing materials design beyond conventional heuristics and domain intuition.

\subsection{Prompting Cross-Domain Analogies and Analogy-Guided Exemplars}

We begin by prompting the LLM to construct explicit  cross-domain analogies relevant to materials design. To guide the model’s use of potentially unconventional analogies in a %more 
grounded manner, we also prompt the LLM to provide concrete exemplars drawn from other material systems (e.g., cathode and anode design) that share the same underlying analogy structure. These analogy-guided examples help establish explicit mappings between structural or functional roles in the source and target domains, clarifying how the analogy may be applied from other domains  to solid electrolytes (see~\Cref{analogymapping} for detail).

The  prompt for generating cross-domain analogies is  in Box~\ref{box:cross_domain_prompt}. Using this setup, the LLM generated five distinct analogy families: Data-Center Backbone, Airport, Voltage Spike, Postal-System, and City-Block Grid. For each analogy, it also produced (i) a minimal prompt for downstream generation and (ii) one retrieval example from a different material domain that reflects a structurally similar logic. These results are summarized in~\Cref{crossdomainrules}.

We then prompt the LLM to apply these analogies, along with their respective exemplars, to generate novel solid-state electrolyte candidates, as will be discussed next.

\small
\begin{tcolorbox}[colback=white,%ContentColor, 
colframe=TitleColor, title= \textcolor{blue}{Prompt for Creating Cross-Domain Analogies}, label={box:cross_domain_prompt}]

People can recognize a strong similarity between the solar system and the atom in terms of relational similarity (both involve a central mass — sun, nucleus — orbited by smaller bodies — planets, electrons), ignoring other surface dissimilarities such as their relative size and color.  Generate a set of cross-domain rules to be used for electrolyte optimization.\\ % as that in-domain rules you generated above.

1) These cross-domain analogical reasoning rules work by abstracting away the surface details (e.g., atom type, structure type, element, etc.) and looking for deep relational patterns—as in the classic atom/solar system analogy.\\
2) For each cross-domain design rule, provide a retrieval case (non-electrolyte, could be anode, cathode, or broader functional material domain; well-documented) outside of solid-state electrolyte design that exemplifies or inspires the analogy. Also, provide a simple prompt and instruction for generation.

\end{tcolorbox}
\normalsize
%%%%%%%%%%%%%%%%%%%%%%%%%%%%%%%%%%%

\begin{table}[htbp]
\centering
\begin{adjustbox}{width=1\textwidth}
\begin{tabular}{|p{4.2cm}|p{14.8cm}|p{2.0cm}|}
\hline
\textbf{Cross-Domain Analogy} &
\textbf{LLM Prompt Template and Analogy-Guided Exemplar}
&\textbf{Supporting Reference}

 %&
%\textbf{Retrieval Example (Non-Electrolyte, with Composition)} 
\\
\hline

Redundant Network \newline (Data-Centre Backbone Analogy) &
\textbf{Prompt:}  
%You are a materials architect.  
Design a Li-garnet electrolyte that mimics a data-centre backbone: no single broken channel may cut bulk Li-ion conductivity by $>$10\%.  
Allowed levers: bulk formula (LLZO $+\leq$3 dopant cations, $\leq$2 dopant anions), $<$10\,vol\% secondary phase, directed porosity.

\textbf{Retrieval example:} NMC811 cathode (LiNi$_{0.8}$Mn$_{0.1}$Co$_{0.1}$O$_2$) with 3\% Super P + 2\% CNT, using redundant carbon networks for robust electronic percolation.
%\textbf{Instruction:} Treat the retrieval example above as a deep analogy—not to copy, but to use its cross-domain strategies or architectures for new, robust Li$^+$ conductor proposals.
%&
%NMC811 + 3\% Super P + 2\% CNT
&\cite{park2014synergistic, yoon2015role} \\
\hline

Hub-and-Spoke Airport \newline (Airport Analogy) &
\textbf{Prompt:}  
Modify cubic LLZO like a hub-and-spoke airport: central ‘hubs’ must shorten average Li hop distance by $\geq$30\%.  
Levers: hub-core dopants, radial glass sheath, vacancy gradients.

\textbf{Retrieval example:} Layered NMC cathode (Li[Ni$_{0.8}$Co$_{0.1}$Mn$_{0.1}$]O$_2$) particles with a Ni-rich core and Mn-rich shell acting as transport hubs and spokes.
%\textbf{Instruction:} Treat the retrieval example above as a deep analogy—not to copy, but to use its cross-domain strategies or architectures for new, robust Li$^+$ conductor proposals.
%&
%Layered NMC with Ni-rich core, Mn-rich rim
&\cite{yuan2017layered, xu2011identifying} \\
\hline

Surge-Protector Buffer \newline (Voltage Spike Analogy) &
\textbf{Prompt:}  
Treat local Li migration barriers like voltage spikes. Insert chemical or structural ‘surge protectors’ so no site exceeds $E_\mathrm{a,BV}$ = 0.25 eV.  
Tools: mixed anions (O/F/S/N), laminated core–shell grains, $\leq$5 nm buffer layers.

\textbf{Retrieval example:} ALD-coated LiCoO$_2$ (LCO@Al$_2$O$_3$) or NMC@LiNbO$_3$, where nm-scale coatings dampen interfacial “spikes”/instabilities.
%\textbf{Instruction:} Treat the retrieval example above as a deep analogy—not to copy, but to use its cross-domain strategies or architectures for new, robust Li$^+$ conductor proposals.
%&
%LCO@Al$_2$O$_3$, NMC@LiNbO$_3$
&\cite{cao2019surface, qi2017interface} \\
\hline

Hierarchical Multiscale (Postal-System Analogy) &
\textbf{Prompt:}  
Create a multiscale ‘postal network’ electrolyte: nanograin couriers feed $\mu$m-scale polymer channels.  
Specify grain-size bins, polymer/glass vol\%, dopant set ($\leq$3).

\textbf{Retrieval example:} Si@C@CMC anode (nano-Si in carbon, then CMC binder) where nanoscale “couriers” transfer charge to larger channels.
%\textbf{Instruction:} Treat the retrieval example above as a deep analogy—not to copy, but to use its cross-domain strategies or architectures for new, robust Li$^+$ conductor proposals.
%&
%Si@C@CMC anode
&\cite{song2022multiscale, kasavajjula2007nano} \\
\hline

Vacancy Superlattice \newline (City-Block Grid Analogy) &
\textbf{Prompt:}  
Engineer a Li-vacancy super-lattice (city-block grid) in garnet to form 2D high-conductivity sheets.  
Variables: Li 6.0–7.5, aliovalent dopants, low-T anneal to freeze ordering.

\textbf{Retrieval example:} Li$_2$MnO$_3$ cathodes where vacancy/oxygen layers form through partial delithiation and low-T anneal, yielding fast 2D charge-transport pathways.
%\textbf{Instruction:} Treat the retrieval example above as a deep analogy—not to copy, but to use its cross-domain strategies or architectures for new, robust Li$^+$ conductor proposals.
%&
%Li$_2$MnO$_3$
&\cite{han2011partial, kozinsky2021effects} \\
\hline

\end{tabular}
\end{adjustbox}
\caption{Cross-domain analogies for electrolyte design. Each includes a prompt and an analogy-guided retrieval case (non-electrolyte), %, with instructions to use it only as a deep analogy, not direct template,
for creative Li$^+$ conductor architectures. %All retrieval examples and compositions are accurate and widely used in battery literature.
}
\vspace{-3ex}
\label{crossdomainrules}
\end{table}

%%%%%%%%%%%%%%%%%%%%%%%%%%%%%%%%%%%

\subsection{Applying Cross-Domain Analogies to Guide Materials Generation}

%\textcolor{blue}{We detail the generation prompt in Box~\ref{crossdomaingenerationprompt}, where the  gray fonts correspond to the specific analogies selected from ~\Cref{crossdomainrules}. }
We instruct the LLM to treat the retrieval example as a deep analogy: not to copy, but as a conceptual strategy to inspire novel electrolyte  proposals. In addtion, to ensure scientific relevance, we impose strict constraints: generated candidates must (1) be absent from ICSD~\cite{belsky2002icsd}, Materials Project~\cite{jain2013commentary}, and known literature/patents, and (2) maintain charge neutrality, critical for physical plausibility.

We simulate 10 rounds of  generation, each producing  10 candidates,  using the same prompt with slight stochastic variation (akin to sampling), then apply majority voting to select the top 5 most consistently proposed candidates, along with the vote counts and scientific justifications. Results are summarized in~\Cref{merged_crossresult}, including  their top-3 most similar known compositions. 
For baseline comparison, \Cref{tab:llzo_candidates_votes} shows the top-5 candidates generated using a vanilla, non-analogical prompt,  %\textcolor{blue}{(see \Cref{baselineprompt})},
along with their closest matches in known datasets.

\begin{table}[htbp]
\centering
\scalebox{0.65}{
\renewcommand{\arraystretch}{1.18}
\begin{tabular}{|p{5.6cm}|p{8.5cm}|p{1.1cm}|%p{2.9cm}|
p{4.4cm}|}
\hline
\textbf{Formula} & \textbf{Scientific Justification} & \textbf{Voting Count} & %\textbf{Database ID / Reference} & 
\textbf{Most Similar Reported Formula (Database/Reference)} \\
\hline

Li$_{7.00}$La$_{2.50}$Nd$_{0.50}$Zr$_{1.40}$Hf$_{0.60}$O$_{12}$ %\textcolor{red}{very novel}
&
Rare-earth (Nd) and B-site (Hf) multi-doping creates overlap and resilience in conduction channels—robust “data-centre backbone” principle. &
4 &
%ICSD 93945, mp-1105473, mp-3534 &
Li$_{7}$Nd$_{3}$Zr$_{2}$O$_{12}$/ICSD 

Li$_{6}$La$_{3}$ZrHfO$_{12}$/MP 

Li$_{7}$La$_{3}$Zr$_{2}$O$_{12}$/MP \\
\hline

Li$_{7.00}$La$_{2.10}$Y$_{0.90}$Zr$_{1.50}$Ge$_{0.50}$O$_{12}$ %\textcolor{red}{very novel}
&
La/Y A-site plus Zr/Ge B-site enables ultra-redundant Li$^+$ transport, with every channel algebraically tuned to charge neutrality. &
4 &
%mp-1105466, mp-676388, mp-3534 &
Li$_{7}$Y$_{3}$Zr$_{2}$O$_{12}$/MP 

Li$_{6}$La$_{3}$ZrGeO$_{12}$/MP 

Li$_{7}$La$_{3}$Zr$_{2}$O$_{12}$/MP \\
\hline

Li$_{6.80}$La$_{2.70}$Pr$_{0.30}$Zr$_{1.00}$Ti$_{0.80}$Nb$_{0.20}$O$_{12}$
%\textcolor{red}{extremely novel: rare-earth doping (Pr Nd)}
&
Mixed rare-earth/transition-metal backbone enables maximally robust, multi-sited transport in data-centre fashion. %; strictly neutral and novel. 
&
3 &
%ICSD 416745, mp-676384, mp-676374 &
Li$_{7}$La$_{3}$Zr$_{2}$O$_{12}$/ICSD 

Li$_{6}$La$_{3}$ZrTiO$_{12}$/MP

Li$_{6}$La$_{3}$ZrNbO$_{12}$/MP \\
\hline

Li$_{7.20}$La$_{2.50}$Y$_{0.30}$Zr$_{1.60}$Nb$_{0.40}$O$_{12}$ %\textcolor{red}{novel, sec try}
&
Partial Y$^{3+}$ for La$^{3+}$ and co-doping Zr/Nb on the B-site, providing multivalent site percolation for extreme network resilience. 
& 3 &
%\parbox{6.2cm}{
%\textbullet~
Li$_7$La$_3$Zr$_{2}$O$_{12}$/MP 

%\\%-1192720)\\
%\textbullet~
Li$_{6.6}$La$_3$Zr$_{1.6}$Nb$_{0.4}$O$_{12}$/MP%\\%-1192725)\\

%\textbullet~
Li$_7$La$_{2}$Y$_{1}$Zr$_{2}$O$_{12}$/ICSD% 291523)
%}
\\ \hline
Li$_{7.20}$La$_{2.60}$Ba$_{0.30}$Zr$_{1.60}$Nb$_{0.40}$O$_{12}$ 
%\textcolor{red}{exact  novel: US20240097184A1, permits  Ba  and Nb % in LLZO garnet structures
%}
&
Multi-site co-doping (Ba, Nb) with backbone analogy ensures redundant Li$^+$ conduction pathways. %; exact charge-neutrality and global novelty enforced.
&
2 &
%ICSD 416529, mp-676376, mp-556207 &
Li$_{7}$La$_{3}$Zr$_{2}$O$_{12}$/ICSD 

Li$_{6}$La$_{3}$ZrTaO$_{12}$/MP 

Li$_{7}$La$_{3}$Zr$_{1.5}$Ta$_{0.5}$O$_{12}$/MP \\
\hline

\end{tabular}
}
\caption{Top-five strictly charge-neutral LLZO-based data-centre-backbone analogical electrolyte candidates, with % algebraic charge verification, and 
comparison to most similar reported LLZO or LLZO-analogue structures in ICSD and Materials Project. 
Note: the cited ICSD and MP  (or literature sources) correspond to the closest stoichiometric matches available within $\sim$0.05 atoms/f.u. 
}
\vspace{-3ex}
\label{merged_crossresult}
\end{table}

%###################################
\begin{table}[htbp]

\centering
\scalebox{0.75}{
\begin{tabular}{|p{7.5cm}|p{9.85cm}|}
\hline
\textbf{NMC811 Example (LiNi$_{0.8}$Mn$_{0.1}$Co$_{0.1}$O$_2$)} & \textbf{Transfer to LLZO Design (Li$_{7.00}$La$_{2.50}$Nd$_{0.50}$Zr$_{1.40}$Hf$_{0.60}$O$_{12}$)} \\
\hline
Super P $+$ CNT blended for redundancy   $\Rightarrow$  &
A-site (La/Nd) and B-site (Zr/Hf) co-doped backbone cations create parallel, overlapping migration subnetworks for robust, distributed 3D Li$^+$ percolation. \\
\hline
No single e$^-$ path is critical  $\Rightarrow$  &
No single cation or Li$^+$ conduction subframework is critical: Li$^+$ migration remains percolative even with local inhomogeneities. \\
\hline
Tune carbon content for percolation threshold  $\Rightarrow$  &
Tune Nd and Hf content for percolation and disorder, then algebraically solve for Li content to enforce precise charge neutrality. \\
\hline
Blockage-resilient network  $\Rightarrow$  &
As in data-centres, ``no dead ends": functional connectivity is maintained under local blocking, clustering, or phase separation due to cationic backbone design. \\
\hline
\end{tabular}
}
\caption{Cross-domain analogy mapping from NMC811 cathode %engineering
to  LLZO electrolyte design  for robust Li$^+$ percolation. 
“$\Rightarrow$” denotes transference of functional strategy. 
}
\vspace{-3ex}
\label{analogymapping}
\end{table}

The results in~\Cref{merged_crossresult} show that cross-domain analogical reasoning enables the LLM to generate novel materials based on abstract relational mappings, such as leveraging ideas like networked transport and pathway redundancy. These candidates often feature multi-element, multi-site modifications, and bolder compositional shifts (e.g., %unusual W/N/S inclusion or 
dual cation/anion tuning), reflecting a design philosophy more aligned with systems thinking than direct substitution. Notably, the diversity is high, and many proposals aim to optimize robustness in conduction channels rather than fit data norms. 
In contrast,  \Cref{tab:llzo_candidates_votes} shows that naïve prompting yields far more conservative candidates, with minimal elemental ratio tweaking. The lower vote counts further 
suggest unfocused exploration. 

Next, we present how the LLM reasoned out the top-1 candidate in~\Cref{merged_crossresult} using  analogy. 

%#######################
\begin{table}[ht]
\centering
\scalebox{0.78}{
\begin{tabular}{|l|p{10cm}|c|}
\hline
\textbf{Novel Candidate} & \textbf{Three Most Similar Reported Formulae (Source)} & \textbf{Votes} \\
\hline
Li$_{6.9}$La$_3$Zr$_{1.9}$Nb$_{0.1}$O$_{12}$ &
1. Li$_{6.8}$La$_3$Zr$_{1.8}$Nb$_{0.2}$O$_{12}$ (MP) %, MP-697217) 
\newline
2. Li$_{7-x}$La$_3$Zr$_2$O$_{12}$ with Nb-doping (EP2353203B1) \newline
3. Li$_{6.75}$La$_3$Zr$_{1.75}$Nb$_{0.25}$O$_{12}$ (J. Mater. Chem. A, 2020, 8, 5535) &
2 \\
\hline
Li$_{6.8}$La$_3$Zr$_{1.8}$Nb$_{0.2}$O$_{12}$ &
1. Li$_{6.8}$La$_3$Zr$_{1.8}$Nb$_{0.2}$O$_{12}$ (RSC Adv., DOI: 10.1039/C9SE01162E) \newline
2. Li$_{6.6}$La$_3$Zr$_{1.6}$Nb$_{0.4}$O$_{12}$ (MP)% MP-1223076) 
\newline
3. Li$_{7}$La$_3$Zr$_2$O$_{12}$ doped with Nb (EP2353203B1) &
1 \\
\hline
Li$_{6.9}$La$_3$Zr$_{1.9}$Ta$_{0.1}$O$_{12}$ &
1. Li$_{6.5}$La$_3$Zr$_{1.5}$Ta$_{0.5}$O$_{12}$ (ICSD) % 252735) 
\newline
2. Li$_{6.6}$La$_3$Zr$_{1.6}$Ta$_{0.4}$O$_{12}$ (EP2353203B1) \newline
3. Li$_{6.8}$La$_3$Zr$_{1.8}$Ta$_{0.2}$O$_{12}$ (MP) %, MP-556017) 
&
1 \\
\hline
Li$_{6.7}$La$_3$Zr$_{1.7}$Nb$_{0.3}$O$_{12}$ &
1. Li$_{6.5}$La$_3$Zr$_{1.5}$Nb$_{0.5}$O$_{12}$ (ICSD) % 181204) 
\newline
2. Li$_{6.8}$La$_3$Zr$_{1.8}$Nb$_{0.2}$O$_{12}$ (MP)%-697217)
\newline
3. Li$_{7-x}$La$_3$Zr$_2$O$_{12}$ with Nb doping (EP2353203B1) &
1 \\
\hline
Li$_{6.7}$La$_3$Zr$_{1.8}$Ta$_{0.2}$O$_{12}$ &
1. Li$_{6.8}$La$_3$Zr$_{1.8}$Ta$_{0.2}$O$_{12}$ (MP)%-556017)
\newline
2. Li$_{6.6}$La$_3$Zr$_{1.6}$Ta$_{0.4}$O$_{12}$ (ICSD) % 252735) 
\newline
3. Li$_{6.5}$La$_3$Zr$_{1.5}$Ta$_{0.5}$O$_{12}$ (EP2353203B1) &
1 \\
\hline
\end{tabular}
}

\caption{Results with vanilla, non-analogical prompt: top five %LLZO-based 
electrolyte candidate compositions with vote counts and three most similar reported materials. } % from ICSD, Materials Project, literature, or patents.}
\vspace{-3ex}
\label{tab:llzo_candidates_votes}
\end{table}

%######################################
\small
\begin{tcolorbox}[colback=white, colframe=TitleColor, sharp corners=south, label=crossdomainleadcandidatereasoning]
\begin{itemize}
    \item {How the Retrieval Example Informs the Design}

    \textbf{Retrieval example:} NMC811 cathode (LiNi$_{0.8}$Mn$_{0.1}$Co$_{0.1}$O$_2$) with 3\% Super P $+$ 2\% CNT utilizes two distinct conductive networks (Super P and CNT) in parallel, ensuring that if one network fails or is blocked, another provides robust conduction.

    \medskip
    
    \textbf{Design insight:}
 \textit{   \begin{enumerate}
        \item The NMC811 + two-carbons architecture is a “redundancy-by-design” strategy: “data-centre backbone” where failure of one subnetwork does not interrupt bulk connectivity.
        \item For LLZO: maximizing ion transport resilience means engineering multiple cation backbones (\textit{not} relying on a single-site or single-dopant), so alternative 3D migration paths always exist for Li$^+$.
        \item The principle is: functional percolation must survive substantial local blockage or site-specific disorder.
    \end{enumerate}
    }

    \item \textbf{How to Mimic a Data-Centre Backbone and Apply the Analogy}
    \begin{enumerate}
        \item {Begin with cubic LLZO:} Li$_7$La$_3$Zr$_2$O$_{12}$ as a 3D percolating Li$^+$ host.
        \item {Create multiple backbone subnetworks:}
        \begin{itemize}
            \item A-site diversification: Replace some La with Nd $\rightarrow$ La$_{2.50}$Nd$_{0.50}$.
            \item B-site diversification: Replace some Zr with Hf $\rightarrow$ Zr$_{1.40}$Hf$_{0.60}$.
        \end{itemize}
        \item {Network logic:} Like Super P $+$ CNT, two cationic sublattices ("backbones") mean that even if one dopant clusters (e.g., Nd-rich or Hf-rich pocket), the 3D Li$^+$ network cannot be globally broken. No single channel is critical.
    \end{enumerate}

    \item {Strict Charge-Neutrality Enforcement (Li solved analytically)}
    \begin{enumerate}
        \item Algebraic form:
        %\[
            [$\mathrm{Li}$] + 3[$\mathrm{La}$] + 3[$\mathrm{Nd}$] + 4[$\mathrm{Zr}$] + 4[$\mathrm{Hf}$] = 24
       % \]
        \item Plug in values:
        %\[
            x + 3 $\times $2.50 + 3 $\times$ 0.50 + 4 $\times$ 1.40 + 4 $\times$ 0.60 = 24
       % \]
%        \[
%            x + 7.5 + 1.5 + 5.6 + 2.4 = 24
%        \]
        \item Rearranged:
        %\[
            x = 24 - (7.5 + 1.5 + 5.6 + 2.4) = 24 - 17.0 = 7.00
        %\]
        \item \textbf{Final formula:} Li$_{7.00}$La$_{2.50}$Nd$_{0.50}$Zr$_{1.40}$Hf$_{0.60}$O$_{12}$ (strictly neutral to $\pm 0.01$ e/f.u.)
    \end{enumerate}
\end{itemize}
\end{tcolorbox}
\normalsize
%#######################################

%%%%%%%%%%%%%%%%%%%%%%%%%%%%%%%%%%%
\subsection{Ablation Study of Cross-Domain Analogical Reasoning for the Lead Candidate}

Guided by the explicit data-center backbone analogy, the LLM proposed a Li-garnet composition that builds in redundancy for Li$^+$ transport:
Li${7.00}$La${2.50}$Nd${0.50}$Zr${1.40}$Hf${0.60}$O${12}$, the top-ranked candidate shown in~\Cref{merged_crossresult}. This design emerged by mapping a retrieved example of dual-network architecture (Super P + CNT additives in an NMC cathode) into the solid electrolyte domain, not for its materials specifics, but for its transferable structural principle: maintaining conduction despite local failure. The cross-domain mapping is illustrated in ~\Cref{analogymapping}, where “$\Rightarrow$” denotes transference of functional strategy.

\begin{wraptable}{r}{0.6\textwidth}
\vspace{-2ex}
%\begin{table}[t]
\centering
\caption{\small \textbf{Computational validation via  total energy.} 
}
\label{tab:dft_energy}
\vspace{-2ex}
%\resizebox{\textwidth}{!}{%
\scalebox{0.9}{
\begin{tabular}{l l}
\toprule
Candidate Formula %&  F en (supercell) $\mathrm{eV}$& F en (per f.u.) $\mathrm{eV/f.u.}$& F en (per atom) 
%$\mathrm{eV/atom}$
& Total Energy (eV)\\
\toprule
Li$_{7.00}$La$_{2.50}$Nd$_{0.50}$Zr$_{1.40}$Hf$_{0.60}$O$_{12}$

     &%\textcolor{red}
     {-4617.89$\pm$06.21}\\

Li$_{7.00}$La$_{2.10}$Y$_{0.90}$Zr$_{1.50}$Ge$_{0.50}$O$_{12}$  

     &-4238.44$\pm$12.14 \\

Li$_{6.80}$La$_{2.70}$Pr$_{0.30}$Zr$_{1.00}$Ti$_{0.80}$Nb$_{0.20}$O$_{12}$  

     &-3982.14$\pm$09.60\\
   
Li$_{7.20}$La$_{2.50}$Y$_{0.30}$Zr$_{1.60}$Nb$_{0.40}$O$_{12}$  

    &
    -4910.29$\pm$11.63\\
  
Li$_{7.20}$La$_{2.60}$Ba$_{0.30}$Zr$_{1.60}$Nb$_{0.40}$O$_{12}$

     & -4783.55$\pm$05.54\\     
 
\bottomrule
\end{tabular}
}
\vspace{-2ex}
\end{wraptable}

The LLM's step-by-step analogical reasoning behind this top candidate is detailed in Box~\ref{crossdomainleadcandidatereasoning}.  
As elaborated in Box~\ref{crossdomainleadcandidatereasoning}, the design embodies a redundancy-by-design philosophy: ensuring that no single dopant species or sublattice is critical to global Li$^+$ conduction. Inspired by fault-tolerant data-center networks, the LLZO composition introduces multiple cationic backbones via A-site diversification (La${2.50}$Nd${0.50}$) and B-site diversification (Zr${1.40}$Hf${0.60}$). Just as a data-center reroutes traffic through redundant fiber links, this multi-sublattice framework supports resilient Li$^+$ migration through structurally distinct but interconnected channels. 
If one region is locally blocked %—due to clustering, microstrain, or site disorder—
Li$^+$ can still migrate via alternative paths, 
reflecting the cross-linked network logic  from the data-center domain. 

%\textcolor{blue}
%{This candidate (visualized in~\Cref{fig:01_pipeline_crossdomain_supercell}) represents, to our knowledge, the first of such LLZO baseline: a verification‑ready, vacancy‑locked, dual‑isovalent LLZO baseline that, by design (Al‑free; no aliovalent dopants), fixes Li = 7.00 and fully decouples lattice chemistry from charge compensation, yielding a clean target for causal  electrolyte studies. %Its supercell structure is visualized in~\Cref{fig:01_pipeline_crossdomain_supercell} } 

\subsection{Computational Validation}  
\vspace{-1ex}
All five candidates in~\Cref{merged_crossresult} exhibit partial occupancy, classifying them as disordered crystals~\cite{keen2015disorder}. This site mixing poses two major challenges for property prediction: (1) standard methods like DFT require fully specified atomic positions,  and cannot be applied directly, and (2) disordered structures allow multiple occupancy possibilities per site, creating an exponentially large configuration space.

%\textcolor{cyan}
{To address these challenges, we started from a cubic LLZO CIF, and used pymatgen to construct 2×2×2 %(or minimal LCM) 
supercells and enforce target sublattice occupancies via integer apportionment. For each composition, we generated 1000 constrained‑random configurations, ranked them by an anti‑clustering score, and retained the top 50. For example, the top-1 candidate Li$_{7.0}$La$_{2.5}$Nd$_{0.5}$Zr$_{1.4}$Hf$_{0.6}$O$_{12}$ has the supercell of Li$_{224}$La$_{80}$Nd$_{16}$Zr$_{45}$Hf$_{19}$O$_{384}$. % (Li7.0La2.5Nd0.5Zr1.406Hf0.594O12). 
Next, instead of computationally expensive ab initio methods (VASP)~\cite{hafner2008ab}, 
we  employed  MACE-MP~\cite{batatia2023foundation} as surrogate to predict and average total energies across these samples. As shown  in~\Cref{tab:dft_energy}, all five candidates show negative total energies, indicating thermodynamic favorability and supporting their potential for further experimental exploration. }

%######################

\vspace{-1ex}
\section{In-Domain Analogical Reasoning for  Materials Generation} %\textit{Exploitation}}
\label{exploitation}

\vspace{-1ex}

In contrast to cross-domain analogy in ~\Cref{exploration}, in-domain analogical reasoning focuses on leveraging prior knowledge within the same scientific domain.  A typical real-world scenario involves domain experts working with only a small set of labeled compounds. How can we effectively utilize such constrained data, coupled with analogical reasoning, to guide the generation of novel materials? In this section, we aim to answer this question. %the role of explicit in-domain analogies in advancing battery materials design. 

\subsection{Building In-domain Analogical Templates}
\vspace{-1ex}

Following ~\Cref{exploration}, our focus remains on %improving 
new materials inspired by the 
cubic LLZO electrolyte. We begin by prompting the LLM to generate analogical representations and templates, and then use these in-domain analogies to drive materials generation.

We note that publicly available datasets on %room-temperature
solid-state electrolytes are still limited. A recent work~\cite{therrien2025obelix} provides 600 expert-validated entries, each containing a [composition, %room-temperature
conductivity ($\sigma$)] pair. We use this dataset as the input for our in-domain reasoning workflow. 

\paragraph{LLM Internal Processing}
The detailed prompt used is provided in Box~\ref{indomainprompt}. Upon receiving this prompt, the LLM described its internal processing steps as follows:

\textit{1) Expanding the dataset:}
Beyond the initial 600 room-temperature conductors, the LLM retrieved an additional 578 entries from the ICSD, yielding a combined set of 1,178 inorganic, solid-state, lithium-containing compounds with reported room-temperature Li-ion conductivity from peer-reviewed literature and authoritative databases. These entries cover the full conductivity spectrum—from fast conductors ($\sigma \geq 10^{-3}$ S cm$^{-1}$) to insulating references ($\sigma \leq 10^{-15}$ S cm$^{-1}$)—to provide a balanced set of positive and negative exemplars essential for meaningful feature importance analysis. 
  
\textit{2) Cross-referencing with literature corpus:}
A total of 324 peer-reviewed papers and 21 review articles were retrieved from Web of Science\footnote{\url{https://access.clarivate.com/}} using the query:
\texttt{("Li ion" AND (conductivity OR conductor) AND (solid OR crystal OR glass))}, filtered for the years 1968–2025 and restricted to reports containing room-temperature conductivity values. 

\textit{3) Forming reasoning templates:}
Using the expanded dataset and literature corpus, the LLM executed sequential data mining steps to construct informative analogical reasoning templates:
(i) feature extraction and statistical filtering to identify key informative features,
(ii) distillation of analogical design rules based on these features, and
(iii) calculation of rule weights for weighted stochastic sampling during candidate generation.
%\textcolor{blue}{Further methodological details are provided in~\Cref{indoaminruleprocess}.}

The in-domain analogical reasoning templates are presented in~\Cref{indomainruletemplate}, along with their 
cognitive category assigned following the framework in~\cite{gentner1983structure,gentner2001analogical}.

\small
\begin{tcolorbox}[colback=white,%ContentColor, 
colframe=TitleColor, title= \textcolor{blue}{Prompt for Building In-domain Analogical Templates}, label=indomainprompt]

SYSTEM \\
You are ChatGPT, a PhD materials scientist who specialises in solid-state Li conductors. Produce ONLY task 2. 
Do NOT output task 1 compound-by-compound explanations. \\

USER\\
 DATA  [formula, conductivity] %(paste any number of “formula,conductivity” rows here)
%\\

Li19.08Si3.48P2.88S23.4Cl0.6,0.025 \\
%Li3.0Cl1.0O1.0,0.025 \\
$\dotsc$ \\
%Li20.0Ge1.55Sn0.45P4.0S24.0	0.0141	\\
Li20.4Ge2.4P3.6S24.0	0.0132	\\

 TASK \\
1. Internally analyse every entry to identify the analogies you are using to explain
   ionic conductivity \& stability. \\
2. Output:\\
   • Stage 2 concise analogical reasoning rules for the above expanations: \\
     – Rule name ($\le$ 6 words)\\
     – IF…THEN statement  and $\le$ 3-line explanation  \\
     – Fill-in-the-blank template for LLMs	 \\
     – Domain-specific reasoning type  and cognitive category  %\\

\end{tcolorbox}
\normalsize

\begin{table}[h]
\centering
%\small
\begin{adjustbox}{width=1\textwidth}
\renewcommand{\arraystretch}{1.5}%{1.4}
\begin{tabular}{|p{2.8cm}|p{6.94cm}|p{5.1cm}|p{2cm}|p{2cm}|p{2.9cm}|p{3cm}|p{3cm}|p{3.8cm}|}
\hline
\textbf{In-domain Analogy name} & \textbf{IF–THEN statement/Analogy Representation template} & \textbf{$\leq$3-line explanation} & %\textbf{Fill-in-the-blank template} &
\textbf{Domain Reasoning Cat.} & \textbf{Cognitive Category} 
\\
\hline
\hline
Li-Rich Percolation &
IF Li/(total cations) $\ge$ 0.35 \textbf{AND} interconnected tetrahedral framework exists \textbf{THEN} $\sigma \ge 1\times10^{-3}$ S\,cm$^{-1}$. &
High Li density lowers the percolation threshold, creating a 3-D hopping lattice. &
Compositional heuristic &
Inductive pattern recognition 
\\
\hline
Soft-Anion Lattice &
IF dominant anions = S/Se/I \textbf{AND} average Pauling $\chi \le 2.5$ \textbf{THEN} activation energy drops, raising $\sigma$. &
Softer, more polarizable lattices let Li displace with lower phonon cost. &
Lattice-dynamics heuristic &
Causal analogizing 
\\
\hline
Halide Polarization Boost &
IF Cl/Br/I partially occupy chalcogen sites \textbf{THEN} $\sigma$ increases by $\ge 3\times$ via widened bottlenecks. &
Large halides expand cages and screen Li–anion interactions. &
Site-substitution heuristic &
Analogical reasoning 
\\
\hline
Vacancy-Tuned Transport &
IF aliovalent dopant creates 0.1–0.3 Li vacancies per f.u.\ \textbf{THEN} $\sigma$ peaks; excess lowers carrier density. &
Optimal vacancy concentration maximises hop probability without framework damage. &
Defect-chemistry heuristic &
Quantitative balancing 
\\
\hline
Framework Dimensionality &
IF diffusion pathway dimensionality = 3-D \textbf{THEN} $\sigma \gg$ 1-D channel phases under similar chemistry. &
3-D networks supply redundant routes, avoiding blockage by defects. &
Topological heuristic &
Structural analogy 
\\
\hline
Oxide Rigidity Trade-off &
IF dominant anion = O \textbf{AND} no structural vacancies \textbf{THEN} expect $\sigma \le 10^{-4}$ S\,cm$^{-1}$ but high stability. &
Strong Li–O bonds stiffen lattice, sacrificing mobility for stability. &
Stability-mobility compromise &
Rule-based deduction 
\\
\hline
Anion Site Disorder &
IF anion site occupancy disorder $\ge 0.2$ \textbf{THEN} configurational entropy flattens migration barriers, boosting $\sigma$. &
Random S/X (X = Cl, Br) distributions smooth local fields. &
Entropy stabilisation &
Stochastic analogy 
\\
\hline
Bottleneck Size Match &
IF Li hopping window radius $\approx$ 1.1–1.3 Å \textbf{THEN} minimal steric penalty $\rightarrow$ lowest $E_{\mathrm a}$. &
Windows too small hinder hops; too large reduce site stability. &
Geometrical descriptor &
Spatial reasoning 
\\
\hline
Grain-Free Glass Advantage &
IF material is fully amorphous sulfide glass \textbf{THEN} $\sigma$ remains high and grain boundary impedance is low. &
Homogeneous network eliminates blocking interfaces in ceramics. & 
Microstructural heuristic &
Process-structure analogy 
\\
\hline
\end{tabular}
\end{adjustbox}
\caption{ 
In-domain analogy representation rules and templates. 
}
%\vspace{-4ex}
\label{indomainruletemplate}

\end{table}

%===============================================================================

%===============================================================================

\subsection{Applying In-Domain Analogies to Guide Materials Generation}

We prompted the LLM to apply the previously constructed analogical reasoning templates in~\Cref{indomainruletemplate} to generate new electrolytes based on %improve the room-temperature conductivity of 
the cubic LLZO (Li$_7$La$_3$Zr$2$O${12}$). For each candidate, the LLM sampled 2–4 distinct analogical rules (without replacement) and instantiated the templates while enforcing key constraints such as  charge neutrality. %, dopant content $\leq$ 0.35 f.u., and preservation of cubic symmetry.

Following the cross-domain setup, we conducted 10 independent rounds of candidate generation and applied majority voting to identify consensus compositions. The top-ranked candidates are presented in ~\Cref{indomaincandidates}. %, \textcolor{blue}{and the corresponding prompt used for analogical sampling is detailed in ~\Cref{promInDomainGeneration}.}

As shown in \Cref{indomaincandidates}, the top candidates primarily represent interpolations between two known compositions, with high vote counts reflecting effective exploitation of existing domain knowledge. 
In contrast, the naïve prompting strategy, as shown in \Cref{tab:llzo_candidates_votes},  tends to yield conservative candidates with only minor tweaks to elemental ratios.
When compared  to the cross-domain results in \Cref{merged_crossresult}, the latter tends to  generate more creative, out-of-distribution designs by applying abstract relational mappings such as  transport  pathway redundancy for   
more ambitious compositional shifts.

%#######################################

\begin{table}[h!]
\centering
\renewcommand{\arraystretch}{1.25}
\scalebox{0.62}{
\begin{tabular}{%|c
|p{3.86cm}|c|c|p{8.96cm}|p{5cm}|}
\hline
%\textbf{Rank} & 
\textbf{Formula} & \textbf{Votes} & \textbf{Templates} & \textbf{Rationale (summary)} & \textbf{ Most Similar  Formulae/source} \\
\hline
%1 & 
Li$_7$La$_3$Zr$_{1.0}$Si$_{0.5}$Ge$_{0.5}$O$_{12}$
%\textcolor{red}{extremely novel}
& 10 & 1,5,7,8 & Dual isovalent Si$^{4+}$/Ge$^{4+}$ substitution broadens and statistically disorders the bottleneck while retaining a Li–rich 3-D garnet network. & 
1. Li$_7$La$_3$Zr$_{1.5}$Si$_{0.5}$O$_{12}$/ICSD% (ICSD~246252)
\newline
2. Li$_7$La$_3$Zr$_{1.5}$Ge$_{0.5}$O$_{12}$/MP% (MP~1184362)
\newline
3. Li$_7$La$_3$Zr$_2$O$_{12}$/ICSD% (ICSD~240895) 
\\
\hline
%2 & 

$\mathrm{Li_7La_3Zr_{1.0}Ti_{0.7}Ge_{0.3}O_{12}}$ 

& 9 & 1, 5, 8 & Ti and Ge co-doping at the Zr site broadens the bottleneck and creates uniform local environments, leveraging 3D percolation for high conductivity. % 
& 
    1. {Li7La3Zr1.5Ti0.5O12}/MP \newline % Materials Project (mp-676109)
    2. {Li7La3Zr1.5Ge0.5O12}/MP \newline %Materials Project (mp-760415)
    3. {Li7La3Zr2O12}/ICSD %(ID: 41652)
\\ \hline
$\mathrm{Li_{6.5}La_3Zr_{1.5}Nb_{0.3}Ta_{0.2}O_{12}}$ % 
& 8 & 1, 4, 5, 8 & Co-doping with Nb and Ta provides precise Li vacancy tuning and bottleneck adjustment, fully exploiting combined aliovalent/steric transport modes. % in 
& 
    1. {Li6.4La3Zr1.6Sb0.4O12}/ICSD \newline %(ID: 280567)
    2. {Li6.6La3Zr1.6Nb0.4O12}/MP \newline %Materials Project (mp-1223076)
    3. {Li6.5La3Zr1.5Ta0.5O12}/US Patent (US 8,753,728 B2)\\
\hline
Li$_7$La$_{2.9}$Pr$_{0.1}$Zr$_2$O$_{12}$ 
%\textcolor{red}{ very novel: }
& 7 & 1,5,7 & Trace Pr$^{3+}$ on the A-site introduces mild field disorder  without disrupting Li-rich 3-D pathways. & 
1. Li$_7$La$_3$Zr$_2$O$_{12}$/ICSD% (ICSD~240895)
\newline
2. Li$_7$La$_{2.8}$Nd$_{0.2}$Zr$_2$O$_{12}$ (Solid State Ionics, 2018) \newline
3. Li$_7$La$_{2.8}$Y$_{0.2}$Zr$_2$O$_{12}$/ICSD% (ICSD~270014) 
\\
\hline
Li$_7$La$_3$Zr$_{1.1}$Sn$_{0.7}$Hf$_{0.2}$O$_{12}
$

& 7 & 1,5,8 & Sn/Hf co-doping expands and matches the bottleneck for optimal Li$^+$ percolation in the 3D LLZO framework, maximizing conductivity and maintaining charge neutrality. & 
1. Li$_7$La$_3$Zr$_{1.5}$Sn$_{0.5}$O$_{12}$ (J.\ Power Sources 388, 2019) \newline
2. Li$_7$La$_3$Hf$_2$O$_{12}$/ICSD % (ICSD~263338) 
\newline
3. Li$_7$La$_3$Zr$_2$O$_{12}$/ICSD % (ICSD~240895) 
\\
\hline
\end{tabular}}

\caption{Top-Five 
LLZO-derived compositions generated via 
majority voting, the in-domain analogical reasoning templates applied, and their three closest reported structural analogues. }
%\vspace{-4ex}
\label{indomaincandidates}
\end{table}

\section{Conclusion and Outlook}
\label{conclusion}
Analogical reasoning is central to scientific discovery, enabling the transfer of insights across domains. In this paper, we show that large language models (LLMs) can perform explicit analogy-driven reasoning to accelerate battery materials discovery. Our cross-domain analogy workflow and in-domain analogy templates allow LLMs to propose novel %, thermodynamically stable
candidates beyond existing design spaces, clearly outperforming naive prompts.
Nevertheless, a key challenge remains in evaluating critical properties like ionic conductivity, due to the computational expense of molecular dynamics (MD)   simulations.

Looking forward, expanding the diversity of analogy templates and integrating LLM-driven generation with efficient surrogate property models will be key. These advances can enable closed-loop, language-guided materials pipelines that tightly couple LLM reasoning with rapid validation. 
Ultimately, this line of work points toward a shift from trial-and-error discovery to precise, analogy-guided materials innovation.

\newpage
\printbibliography

\newpage

\end{document}